\documentclass[]{fairmeta}

\usepackage[utf8]{inputenc} %
\usepackage[T1]{fontenc}    %
\usepackage{hyperref}       %
\usepackage{url}            %
\usepackage{booktabs}       %
\usepackage{amsfonts}       %
\usepackage{nicefrac}       %
\usepackage{microtype}      %
\usepackage{xcolor}         %
\usepackage{multirow}
\usepackage{titlesec}

\titlespacing*{\section}{0pt}{5pt}{3pt}
\titlespacing*{\subsection}{0pt}{4pt}{2pt}
\usepackage[most]{tcolorbox}
\newtcolorbox{prompt}[1]{
    enhanced,
    left=4mm,
    right=4mm,
    top=2mm,
    bottom=2mm,
    boxsep=0mm,
    rounded corners,
    title=#1,
    fontupper=\footnotesize\linespread{0.9}\fontfamily{lmr}\selectfont,
    }

\definecolor{light-gray}{gray}{0.9}
\definecolor{comm}{gray}{0.5}
\definecolor{denim}{rgb}{0.08, 0.38, 0.74}
\usepackage{enumitem}
\hypersetup{
  colorlinks   = true, %
  urlcolor     = denim, %
  linkcolor    = denim, %
  citecolor   =  denim,%
}

\usepackage{amsmath}
\usepackage{amssymb}
\newcommand{\pic}{\pi_\theta}
\newcommand{\piref}{\pi_\text{ref}}
\newcommand{\piold}{\pi_{\theta_\text{old}}}

\usepackage{subcaption}

\title{Bridging Offline and Online\\Reinforcement Learning for LLMs}

\author[1,*]{Jack Lanchantin}
\author[1,2,*]{Angelica Chen}
\author[1]{Janice Lan}
\author[1]{Xian Li}
\author[1]{Swarnadeep Saha}
\author[1]{Tianlu Wang}
\author[1]{Jing Xu}
\author[1]{Ping Yu}
\author[1,2]{Weizhe Yuan}
\author[1,2]{Jason Weston}
\author[1,\dagger]{Sainbayar Sukhbaatar}
\author[1,\dagger]{Ilia Kulikov}

\affiliation[1]{FAIR at Meta}
\affiliation[2]{NYU}

\contribution[*]{Joint first author}
\contribution[\dagger]{Joint last author}

\abstract{We investigate the effectiveness of reinforcement learning  methods for finetuning large language models when transitioning from offline to semi-online to fully online regimes for both verifiable and non-verifiable tasks. Our experiments cover training on verifiable math as well as non-verifiable instruction following with a set of benchmark evaluations for both. Across these settings, we extensively compare online and semi-online Direct Preference Optimization and Group Reward Policy Optimization objectives, and surprisingly find similar performance and convergence between these variants, which all strongly outperform offline methods. We provide a detailed analysis of the training dynamics and hyperparameter selection strategies to achieve optimal results. Finally, we show that multi-tasking with verifiable and non-verifiable rewards jointly yields improved performance across both task types.}

\date{\today}
\correspondence{Jack Lanchantin \email{jacklanchantin@meta.com}, Ilia Kulikov \email{kulikov@meta.com}}

\begin{document}

\maketitle

\section{Introduction}

Large Language Models (LLMs) have demonstrated remarkable capabilities on a wide variety of tasks spanning open ended instruction following to rigid mathematical reasoning \citep{dubey2024llama,shao2024deepseekmath}. A key ingredient for these capabilities is the ``post-training'' or alignment stage where a base language model is shaped for specific tasks. During this phase the model is fine-tuned via Reinforcement Learning (RL) to optimize for human preferences or verifiable rewards. 
The former is suitable for open-ended generations and takes advantage of a reward model during training to reduce reliance on human annotators.
The latter is used for math, code, and other less open-ended questions where the correctness of the answer can be directly verified with a boolean score typically by matching against existing labels. 

For the optimization method itself, several candidates are commonly considered.
When learning from preference labels, Direct Preference Optimization (DPO) \citep{rafailov2024direct} has emerged as a powerful algorithm and became a popular choice for open-ended tasks due to its simplistic offline training \citep{xu2024dpo}.
It can be used with verifiable rewards \citep{pang2024iterative} or with reward models \citep{xu2023some}. DPO can also be used in a semi-online (iterative) fashion \citep{pang2024iterative, yuan2024self}.
More recently, however, Group Relative Policy Optimization (GRPO) \citep{shao2024deepseekmath} has become widely used for fine-tuning LLMs in an online fashion for its success in training thinking LLMs \citep{guo2025deepseek}.
GRPO is based on a popular RL algorithm PPO \citep{schulman2017proximal} which belongs to a class of online training methods that try to estimate the gradient of the reward signal.

\begin{figure}
    \centering
    \includegraphics[width=1.0\linewidth]{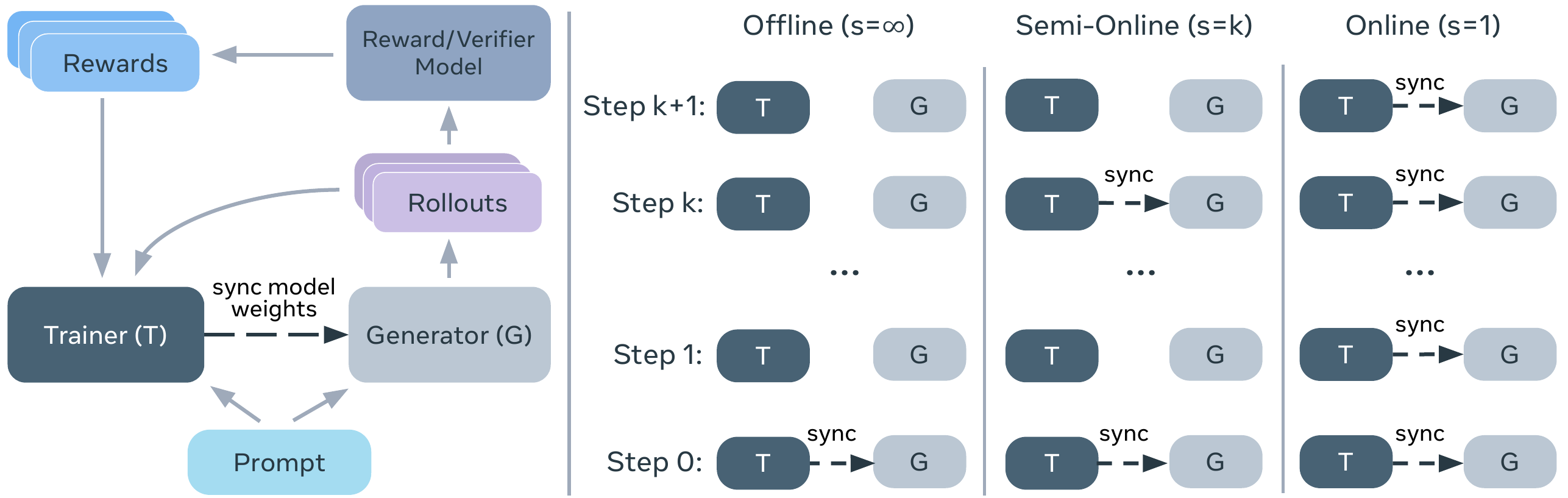}
    \caption{\textbf{(left):} Visualization of a single training step within our training pipeline, which can be used for any training objective such as DPO or GRPO. 
    Syncing the weights allows rollout responses to be generated from the most recent model.
    \textbf{(right):} Progression from offline to online training, showing when model weight synchronizations occur at different train steps. Offline training only syncs before training starts, whereas online training syncs at every step.
    }
\label{fig:offline_to_online_training}
\end{figure}

While recent models have achieved impressive benchmark results, the relative importance of various offline to online training approaches and their generalization performance across different tasks remains poorly understood.
In this paper, we systematically explore the effectiveness of LLM post-training methods in different training setups by bridging the gap between offline and online methods. Specifically, we study offline, semi-online, and online configurations, across both verifiable and non-verifiable tasks, as depicted in \autoref{fig:offline_to_online_training}. By examining the transition from offline to online training, i.e., by altering the speed of periodic model syncing, we aim to understand how these methods can be optimized for improved performance and efficiency on any task. Our investigation focuses on two key aspects: the comparative effectiveness of semi-online or fully online training over offline training and the relative performance of DPO and GRPO objectives across verifiable and non-verifiable tasks.

Based on our experimental results, our contributions are as follows.
First, we show that standard DPO lags behind other training regimes significantly, likely due to its offline nature.
In contrast, online DPO achieves comparable performance to online GRPO, but more surprisingly so does  semi-online DPO.
We make several recommendations for making such training more stable.
The efficiency gains of the semi-online variants opens up an interesting question of whether fully online RL is the only approach for post-training LLMs.
Finally, we investigate the performance of joint optimization of verifiable tasks with rule-based rewards and non-verifiable tasks with reward models.
We find that this results in improved average results across all tasks compared to the baseline of optimizing only on one objective or the other, as expected, and observe that it improves non-verifiable evaluations compared to only training on non-verifiable rewards.

\section{LLM alignment algorithms}

LLM alignment or post-training is performed after the initial pre-training stage. The de-facto task definition for LLM alignment is an instruction following task where the model input specifies instruction and auxiliary task constraints, and a (typically human-written) response is used as the target. Due to its practical scalability, supervised fine-tuning (SFT) was initially the most common approach to post-train using high-quality instruction following data \citep{touvron2023llama1, touvron2023llama, zhou2023lima}. 
Reinforcement Learning from Human Feedback (RLHF) was proposed before the rise of assistant-like LLMs \citep{ziegler2019fine}, and it it was only relatively recently that it was used to outperform SFT methods \citep{ouyang2022training}. This was made possible by  instruction following datasets being annotated with a set of responses and human preference labels corresponding to each response, allowing the training of reward models. Initial RLHF models were finetuned using Proximal Policy Optimization (PPO) \citep{schulman2017proximal}. More recently, Direct Preference Optimization \citep{rafailov2023direct} and Group Relative Policy Optimization \citep{shao2024deepseekmath} have become the gold standard finetuning methods for aligning language models. We detail these methods in the following subsections as they provide the basis for our experiments.

\subsection{Group Relative Policy Optimization (GRPO)}
\label{sec:grpo}
GRPO \citep{shao2024deepseekmath} is based on the PPO \citep{schulman2017proximal} algorithm, an on-policy policy-gradient method (\autoref{sec:extended_background_alignment}). While PPO learns from a single sample, which makes it generally applicable, GRPO leverages the fact that we can sample a group of responses {\small$G=\{y^1, \ldots, y^N\}$} for any given prompt $x$.
This allows us to approximate a relative advantage of each response by {\small$A(y^i|x) = r(y^i|x) - \sum_{y_j \in G} r(y_j|x)/N$}.
{\small
\begin{align}
\mathcal{L}_\text{GRPO} = -\mathbb{E}_{G \sim\piold} \left[ \sum_{y^i \in G} \sum_t \min \left\{ \frac{\pic(y_t|x,y_{<t})}{\piold(y_t|x,y_{<t})} A(y^i), \text{clip}_\epsilon\left( \frac{\pic(y_t|x,y_{<t})}{\piold(y_t|x,y_{<t})} \right) A(y^i) \right\}  \right].
\end{align}
}We do not normalize by length like in \citet{shao2024deepseekmath} because \citet{liu2025understanding} showed that it can lead to biased optimization, and it is not in the original PPO loss. There is an additional KL term in the loss that we omitted here for brevity. 

The main advantage of PPO is that it allows for a small amount of off-policy learning by sampling from an outdated policy $\piold$. This enables efficient training by performing multiple updates on the same batch of generations.
The loss uses per-step importance sampling, which is more stable than sequence level importance sampling, and proven to be unbiased \citep{schulman2015trust}. The proof relies on the fact that the advantage term is for a single step of the policy
{\small
\begin{align}
    A_\pi(y_t) = r(y_t) + V_\pi(y_t) - V_\pi(y_{t+1}) .
\end{align}
}However, the advantage term of GRPO is at the sequence level, so we cannot reuse the same proof and the paper does not provide its own proof \citep{shao2024deepseekmath}.
Therefore, we restrict our experiments to a purely on-policy setup without importance sampling when using the GRPO loss.

\subsection{Direct Preference Optimization (DPO)}
DPO \citep{rafailov2024direct} is an offline alignment algorithm that is derived from RLHF \citep{ziegler2019fine,ouyang2022training} and designed to learn from preference labels $y_c \succ y_r$ where response $y_c$ is deemed better than $y_r$ for prompt $x$. The DPO loss is as follows (see \autoref{sec:extended_background_alignment} for derivation):
{\small
\begin{align}
\mathcal{L}_\text{DPO}= -\log \sigma \left( \beta \log \frac{\pi(y_c|x)}{\piref(y_c|x)} - \beta \log \frac{\pi(y_r|x)}{\piref(y_r|x)} \right) .
\end{align}
}Unlike PPO or GRPO that directly optimize the reward with noisy estimates based on a single sample, DPO optimizes the relation between two samples to match the optimal setup, which can be calculated from data without noise.
While this reduced training noise is an advantage, DPO lacks a theoretical guarantee on how a decrease in loss increases the expected reward.
Another advantage of DPO, however, is that it does not rely on how the samples are generated, making it appealing for off-policy settings where responses are generated by another model.

\subsection{Semi-online Optimization}
As described above, GRPO is an on-policy algorithm that requires samples to be generated from the current policy, whereas DPO can learn from off-policy samples (\autoref{fig:offline_to_online_training}).
Therefore, the GRPO training pipeline must be online -- \emph{i.e.}, the generations and model updates must be synchronous. 
DPO, on the other hand, was designed for a purely offline setup where we can generate training responses beforehand and train with the DPO loss on these pre-generated responses.
However, it is also possible to perform multiple iterations of DPO where one trains on the entire dataset at each iteration, and then generates a new set of responses using the model from the previous iteration. Iterative DPO often offers performance boosts over offline DPO \citep{xu2023some,yuan2024self,chen2024self}.

In our work we consider a {\em semi-online} DPO setup where the generation model parameters are synchronized with the training model parameters only periodically, but potentially much more often than in the iterative setting just described. 
Let $s$ be a number of parameter update steps performed between each synchronization.
Decreasing $s$ will make it more online, and eventually become purely online at $s=1$ when responses are generated using the latest model parameters.
In our experiments, we bridge the gap between offline and online training by controlling $s$ to see its effect on downstream performance. 
The advantage of reducing $s$ lies in computation efficiency where responses can be generated in an embarrassingly parallel way.

While PPO can also be run in a slight off-policy setup thanks to its importance sampling adjustment, it is an inherently on-policy algorithm and uses clipping to limit the importance sampling ratio. In practice PPO is often limited to several update steps before synchronizing the generator with the current model.
The GRPO paper \citep{shao2024deepseekmath} does not mention if more than one update is performed between synchronizations\footnote{It mentions ``The policy model only has a single
update following each exploration stage.'' which could mean only one update is performed between synchronizations, making it fully online.}, and most open-source implementations use a pure-online setup.
As mentioned in \autoref{sec:grpo}, the off-policy update with GRPO lacks theoretical clarity and is not well studied, so we leave it to future work.

\section{Experimental Setup}

We study the effectiveness of post-training along three main axes: the training recipe (offline, semi-online, online), algorithm (DPO, GRPO), and tasks (non-verifiable, verifiable).

\textbf{Semi-online configurations\ } We analyze how the update rate impacts training performance and stability. 
As mentioned before, after every $s$ model weight updates, the generation model is synchronized to match the current model.
For both tasks, we compare offline DPO ($s=\infty$), online DPO and GRPO ($s=1$), and two semi-online DPO settings that periodically synchronize the generation model ($s\in [5,10,100]$). 
In DPO, we either keep $\pi_\text{ref}$ fixed, or update along with the generator model (see Appendix \autoref{tab:training_hyperparameters} for specific implementations).

\textbf{Hyperparameter settings\ }
For all tasks, we initialize model parameters using the \texttt{Llama-3.1-8B-Instruct} model. During training, we use the default sampling parameters (temperature=$1.0$, top-p=$1.0$) to generate exploration rollouts. Other hyper-parameters such as loss configuration, learning rate, gradient clipping, and optimizer settings differ based on the task we train on and are provided in Appendix \autoref{tab:training_hyperparameters}.

\textbf{Training implementation details\ }
We train all models using the \texttt{fairseq2} library \citep{balioglu2023fairseq2}, where model inference is performed with the \texttt{vllm} library \citep{kwon2023efficient}. Our main design goal is to create a flexible and modular framework that can easily change policy models, reward models, training algorithms, and datasets. At the same time, we set forth the objective of fast sequence generations for online optimization algorithms such as online DPO and GRPO. 
We run all experiments using 32 NVIDIA H200 GPUs for training workers and 8 H200 GPUs for inference workers (16 for combined task training). 
We provide further technical details about the online recipe design in \autoref{sec:design_details}.

\subsection{Non-verifiable instruction following}

\textbf{Task\ } Instruction following is an umbrella task that can represent both verifiable and non-verifiable types of questions. Here,  we focus on the distribution of problems that users typically ask LLM assistants. Specifically, we rely on the WildChat-1M dataset \citep{zhao2024wildchat}, which is a collection of 1 million user interactions with ChatGPT. We randomly sample user prompts from the subset of first-turn interactions from the dataset. %
The prompt template we used for a given instruction is given in Appendix \autoref{fig:non-verifiable_prompt}. 

\textbf{Reward\ } The non-verifiable nature of this task, meaning that there is no (unique) reference answer, requires us to employ a reward model that can estimate the quality of the model response given the user input. We use the open-source LLM-based reward model \texttt{Athene-RM-8B} \citep{Athene2024}, which is experimentally validated as one of the best models to use for preference ranking \citep{frick2024evaluaterewardmodelsrlhf}. Athene-RM-8B generates a scalar score for an input-response pair. This allows us to either use the raw response scores as rewards in GRPO, or, in the case of DPO, to rank the responses and create preference pairs of chosen and rejected responses corresponding to the highest and lowest scores, respectively\footnote{While other methods exist to create preference pairs from scalar rewards \citep{lambert2024t}, we choose the best-vs-worst due to its simplicity and stability \citep{yuan2024self, xu2023some, pace2024west}.}.

\textbf{Evaluation\ } For evaluation of the helpfulness and quality of responses, we use AlpacaEval 2.0 \citep{alpaca_eval, dubois2024length} and Arena-Hard \citep{li2024crowdsourced,arenahard2024} 
which are robust instruction following benchmarks that have a high correlation with user preferences. We evaluate using two judges: GPT-4-1106 and GPT-4o.  
We use the decoding temperature $0.6$ and the top-p $0.9$ to generate predictions, which are aligned with the commonly used values of the seed model we use in this work. For training, we use 1,000 WildChat prompts for 1,500 steps. We select the best model checkpoint based on the highest length-normalized Athene-RM-8B rewards on a heldout set of 470 examples: 253 validation examples from \cite{li2023self} and 218 Evol-Test set examples from \cite{xu2023wizardlm}, with prompts that overlap with AlpacaEval 2.0 removed.

\subsection{Verifiable math problems}

\textbf{Verifier\ } Math problems featuring a reference answer together with the input problem have become a standard in the verifiable training setup \citep{lambert2024t, guo2025deepseek}. The core component behind such setup is a robust verifier that can match the predicted answer with the reference one. Some mathematical problems might have multiple written forms of the correct answer e.g., ${2}/{4} = 0.5$ and ${2}/{4} = {1}/{2}$.
As such, we use the open-source verification toolkit Math-Verify\footnote{https://github.com/huggingface/Math-Verify} instead of exact match verification. 
The template LLM prompt for these tasks is given in Appendix \autoref{fig:verifiable_prompt}. The template asks the model to ``reason and give a final answer to the problem'' but does explicitly ask for a separate thinking component \citep{guo2025deepseek} so that it is closer to user instructions from the non-verifiable task.

\textbf{Reward\ } Using the verifier, we obtain binary rewards for each prompt-response pair. 
For DPO, preference pair selection involves randomly picking the chosen response from the pool of correct predictions, and the rejected response from the pool of incorrect predictions. Prompts that are either too easy or complicated can result in pools where all predictions are correct or incorrect, so we cannot form a valid preference pair. In this case we skip this prompt from the current training step.
It is similar in the GRPO loss because all the advantages will be zero.

\textbf{Data\ } We rely on the NuminaMath dataset \citep{li2024numinamath} to collect training problems and reference answer pairs. During data selection, we filter out problems that require generating a proof, multiple choice questions, and synthetic data, including the Orca math, synthetic AMC, and synthetic math subsets. The proof questions are non-trivial to verify using answer matching, the multiple-choice questions incentivize the model to predict any answer from the given options without generating a useful rationale, and the synthetic data may have incorrect answers. After filtering, we end up with a diverse set of 261,440 math problems from which we select $1980$ problems each for our held-out validation and test sets. 

\textbf{Evaluation\ } We evaluate using Math500 \citep{hendrycks2021measuring, lightman2023lets}, AMC23,
and the NuminaMath test set. We use temperature $0.6$ and top-p $0.9$ to generate predictions. For each problem we generate $N=50$ solutions and report the average accuracy as well as the standard error.

\begin{table}[t]
  \centering
\caption{\textbf{Verifiable Task Evaluations}. Test accuracy (std error) for Math500, NuminaMath, and AMC23. 
Standard error is computed over $N=50$ random seeds. We find that all semi-online and fully online methods significantly outperform the seed model and offline training, with semi-online DPO, online DPO and GRPO all performing similarly.}
\label{tab:model_performance_verifiable}
 \resizebox{0.65\textwidth}{!}{
  \begin{tabular}{@{}l c c c@{}}
    \toprule
    \textbf{Training method}                 & \textbf{Math500}         & \textbf{NuminaMath}    & \textbf{AMC23}           \\
    \midrule
    Seed (\texttt{Llama-3.1-8B-Instruct}) & 47.4 (1.6)   & 33.9 (0.6)   & 23.7 (5.2)   \\
    Offline DPO $(s=\infty)$          & 53.7 (1.6)   & 36.4 (0.6)   & 28.8 (7.0)   \\
    \midrule
    Semi-online DPO $(s=100)$   & \textbf{58.9} (1.2)  & 39.3 (0.4)   & \textbf{35.1} (5.3)   \\
    Semi-online DPO $(s=10)$    & 57.2 (1.1)   & 39.4 (0.5)   & 31.4 (4.3)   \\
    \midrule
    Online DPO $(s=1)$            & 58.7 (1.2)   & \textbf{39.6} (0.5)   & 32.9 (5.2)   \\ 
    GRPO              & 58.1 (1.3)   & 38.8 (0.5)   & 33.6 (5.1)   \\
    \bottomrule
  \end{tabular}
  }
\end{table}

\subsection{Combining verifiable and non-verifiable tasks}

\textbf{Skills generalization\ } While many recent works focus on improving reasoning within specific domains (e.g. verifiable math) \citep{lambert2024t, guo2025deepseek}, we ultimately want models to perform well on the whole range of tasks. Previous works indicate that using a reward model based on human preferences can lead to reward hacking and poor performance on verifiable tasks \citep{gao2023scaling, guo2025deepseek}. We therefore are motivated to study overall performance when only training on one type of reward, and when combining both verifiable and non-verifiable rewards to train a single model. That is, we will use the verifiable rewards for verifiable tasks, and the reward model rewards for non-verifiable tasks. The integration of both types of rewards into a unified training run presents a robust test for the ability of reinforcement optimization generalization. In doing so, we aim to demonstrate two capabilities. First, that we can successfully combine different reward types into a single training run. Second, that the fine-tuned model is both verifiably accurate in definitive math problems, as well as highly coherent and helpful in open ended instruction following problems.

\textbf{Data\ } We consider 3 scenarios: further finetuning a Wildchat Online-DPO-finetuned checkpoint with 100k NuminaMath samples, further finetuning a NuminaMath Online-DPO-finetuned checkpoint with 100k WildChat samples, and finetuning a \texttt{Llama-3.1-8B-Instruct} seed model with both WildChat and NuminaMath data. In the last setting, we use 100k NuminaMath prompts and 50k WildChat prompts (since roughly half of the NuminaMath samples are skipped in each batch), and combine samples from both into a single batch. This mixes both verifiable (binary) and non-verifiable (scalar) rewards at each training step. We select checkpoints based on the highest value when averaging length-normalized non-verifiable reward and verifiable reward on their respective validation sets. For all combination training runs, we train for a maximum of 15k steps.

\section{Results}

\subsection{Main Results}

\textbf{Verifiable math\ }
~\autoref{tab:model_performance_verifiable} shows math evaluation results for the different training regimes on the NuminaMath training set. The offline DPO training improves performance across all benchmarks compared to the seed model. However, we see substantial gains when training in online, or semi-online regimes.
We observe several important trends. First, online and semi-online trained models $(s\geq1)$ all outperform the offline DPO model $(s=\infty)$ by a wide margin. This highlights the limitation of offline training and the importance of training on responses generated by an updated model.
Second, we notice the effectiveness of operating in a semi-online setting with $(s>1)$ for DPO, which performs very similarly to completely online DPO $(s=1)$. 
This is an important finding indicating that pure online training might not be necessary.
We find that online DPO marginally outperforms GRPO. Lastly, we experiment with different numbers of responses in GRPO and report results in Appendix \autoref{tab:grpo_n_ablation}, where scaling it beyond $8$ did not boost performance further.

\begin{table}[]
    \centering
    \caption{\textbf{Non-Verifiable Task Evaluations}. 
    We show winrate with standard error for length-controlled AlpacaEval, and ArenaHard scores with 95\% confidence intervals. Similar to verifiable tasks, both semi-online and online DPO show the best performance, closely followed by GRPO. We show results using two judges: GPT-4-1106 and GPT-4o. While GPT-4o gives overall lower winrates, we see general relative agreement between the two judges.}
    \label{tab:model_performance_nonverifiable}
    \resizebox{1.0\textwidth}{!}{
    \begin{tabular}{lccccc}
        \toprule
        &   \multicolumn{2}{c}{\textbf{AlpacaEval LC Winrate}} & \multicolumn{2}{c}{\textbf{ArenaHard Score}} \\
    \cmidrule(lr){2-3} \cmidrule(lr){4-5}
     \raisebox{1ex}[0pt]{\textbf{Training Method}} &  {\small{\textbf{GPT-4-1106 Judge}}} &  {\small{\textbf{GPT-4o Judge}}} &   {\small{\textbf{GPT-4-1106 Judge}}} &  {\small{\textbf{GPT-4o Judge}}}  \\
        \midrule
        Seed (\texttt{Llama-3.1-8B-Instruct}) & 27.3 (1.3) & 21.3 (-2.2, 1.7) & 32.0 (1.55)   & 27.8 (-2.1, 1.8) \\
        Offline DPO $(s=\infty)$     & 53.2 (1.5) & 38.3 (-2.8, 2.2)  & 39.4 (1.68) & 38.2 (-2.1, 2.9)\\
        \midrule
        Semi-online DPO $(s=10)$     & 81.6 (1.0) & 59.4  (-1.6, 1.4) & 61.1 (1.62) & 43.0 (-1.6, 1.6)\\
        Semi-online DPO $(s=5)$      & 78.7 (1.2) &  \textbf{60.7} (-1.9, 2.4) & 58.5 (1.49) & 49.8 (-2.2, 2.2)\\
        \midrule
        Online DPO $(s=1)$           & \textbf{83.1} (1.0) & 60.1 (-1.8, 1.6)  & \textbf{62.8} (1.53)  & 50.4 (-1.7, 1.9)\\
        GRPO                         & 75.2 (1.2) & 55.0 (-1.7, 1.8)  & 59.1 (1.40) & \textbf{54.3} (-1.5, 1.6)\\
        \bottomrule
    \end{tabular}
    }
    
\end{table}

\textbf{Non-verifiable instruction following\ }
\autoref{tab:model_performance_nonverifiable} compares the performance of different models training on WildChat prompts with the \texttt{Athene-RM-8B} reward model. We show AlpacaEval-2.0 Length-Controlled (LC) winrates and ArenaHard scores.
We observe improvements over the baseline seed model in all training regimes: offline, semi-online, and online. However, again semi-online and online methods significantly outperform the offline DPO results. For example, averaged across both judges, Online DPO results in a 56.6\% increase in AlpacaEval LC winrate and 45.6\% increase in ArenaHard score compared to the commonly used offline DPO. 

Similar to the verifiable task setting, online DPO results in slightly higher performance compared to GRPO. Hence both settings emphasize the importance of online and semi-online training methods compared to offline. For semi-online DPO, we test smaller semi-online synchronization step sizes $s = \{5, 10\}$ because 32 steps is already a full data epoch, and we find $s=100$ to be too unstable with our non-verifiable hyperparameters. We find similar performance between semi-online and online, reiterating the effectiveness of sync step sizes that we observed in the verifiable task. While it is possible that there is some reward hacking with the \texttt{Athene-RM-8B} reward model via response length (see~\autoref{sec:additional_experiments}), our results demonstrate robust performance on two commonly used instruction following benchmarks that are highly correlated with human preferences and control for common reward hacking such as length and style.

\textbf{Combining verifiable and non-verifiable\ }
Finally, we analyze the effectiveness of training a model with both verifiable and non-verifiable tasks in the training set.
Given the strong performance results in the individual verifiable and non-verifiable tasks, and due to computational resource constraints, we only consider online DPO training in this setting. 
\autoref{tab:combination_results} shows the results of the combined dataset models compared to training on individual verifiable or non-verifiable tasks. 
First, we see that the ``cross'' task performance, training on only verifiable and testing on non-verifiable or vice versa, results in either a decrease in performance or marginal improvement compared to the seed \texttt{Llama-3.1-8B-Instruct} model, i.e. there is no transfer. 
However, we observe significant improvements on non-verifiable tasks when starting from either a WildChat or NuminaMath checkpoint and finetuning on the opposite training set. Notably, even when starting from a checkpoint trained on 1k WildChat and finetuning on 100k NuminaMath samples, we still see gains on non-verifiable evals.  We hypothesize that since AlpacaEval and ArenaHard contain ``verifiable'' prompts such as math and code, it is critical to incorporate some verifiable signal during training as opposed to only using an LLM-based reward signal. When starting from the NuminaMath checkpoint and finetuning on WildChat, we see a significant decrease in math performance as the model starts to optimize for the LLM-based reward.
Lastly, we see performance gains across both verifiable and non-verifiable tasks when starting from the seed model and training on both reward signals. Performance is comparable to training on each task individually, with slight improvements in the non-verifiable evaluations, demonstrating that not only is it possible to combine rewards during training, it can also help improve performance in certain tasks.

\begin{table}[t]
    \centering
    \caption{\textbf{Combined Verifiable + Non-Verifiable Evaluations}. We first show the ``cross'' task evaluations, when training a model on either the NuminaMath (NM) task only or the WildChat (WC) task only, where we see poor cross task transfer. We then show three separate models trained on both task rewards either starting from a checkpoint trained on the opposite task, or training on both at once. 
    We observe better results with the combined task models across all four datasets than any individual-task model. We show accuracy with standard error for MATH500 and AMC23, winrate with standard error for length-controlled AlpacaEval, and ArenaHard scores with 95\% confidence intervals.}
    \resizebox{1.0\linewidth}{!}{%
    \begin{tabular}{lllcccc}
    \toprule
    & & &  \multicolumn{2}{c}{\textbf{Verifiable}} & \multicolumn{2}{c}{\textbf{Non-verifiable (\small{GPT-4o Judge})}} \\
    \cmidrule(lr){4-5} \cmidrule(lr){6-7}
     \raisebox{1ex}[0pt]{\textbf{Seed}} & \raisebox{1ex}[0pt]{\textbf{Training}} & \raisebox{1ex}[0pt]{\textbf{Dataset}} &  {\textbf{MATH500}} &  {\textbf{AMC23}}  & \textbf{{AlpacaEval LC}} & \textbf{{ArenaHard}}  \\
        \midrule
        Llama-3.1-70B-Instr. & \multicolumn{1}{c}{-} & \multicolumn{1}{c}{-} & 67.2 (1.4) & 46.6 (4.3)  & 43.5 (1.6)  & 56.4 (-2.3, 2.4) \\
        \midrule
        Llama-3.1-8B-Instr. & \multicolumn{1}{c}{-} & \multicolumn{1}{c}{-} & 47.4 (1.6) & 23.7 (5.2) &  32.0  (1.6) & 27.8 (-2.1, 1.8)  \\
        Llama-3.1-8B-Instr. & Online DPO & NM only & \textbf{58.7} (1.2) & \textbf{32.9} (5.2) & 36.2 (1.6) & 34.9 (-2.2, 2.2) \\
        Llama-3.1-8B-Instr. & Online DPO & WC only & 35.0 (1.6) & 15.0 (4.3) & 62.8 (1.5) & 50.4 (-1.7, 1.9) \\
        \midrule
        WC-Checkpoint & Online DPO & NM only  & 54.7 (1.2) & 30.4 (5.5)  &  71.9 (1.5) & \textbf{62.3} (-2.1, 2.4) \\
        NM-Checkpoint & Online DPO & WC only & 33.3 (1.6) & 13.2 (4.9)  &  \textbf{78.8} (1.6) & 61.2 (-2.0, 2.3)  \\
        Llama-3.1-8B-Instr. & Online DPO & NM+WC  & 57.3 (1.4) & 31.7 (6.0)  &  65.6 (1.3) & 57.1 (-2.3, 3.0)  \\
        \bottomrule
    \end{tabular}
    }
    
    \label{tab:combination_results}
\end{table}

\begin{figure}[t]
    \centering
    \includegraphics[width=\linewidth]{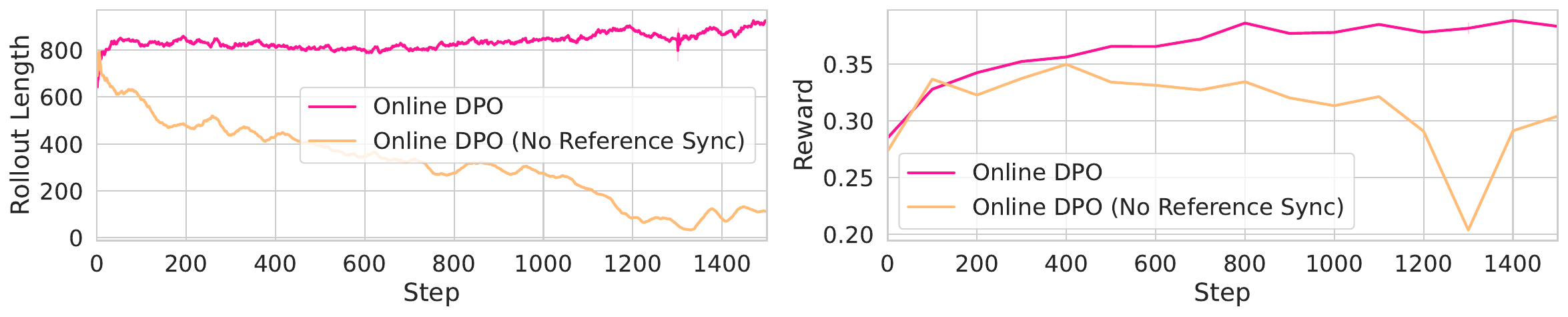}
    \caption{Without syncing the reference model, response lengths of online DPO collapse when trained on verifiable tasks (left). This length collapse is also correlated with lower validation reward (right).}
    \label{fig:length_collapse}
\end{figure}

\begin{figure}[t]
    \centering
    \includegraphics[width=\linewidth]{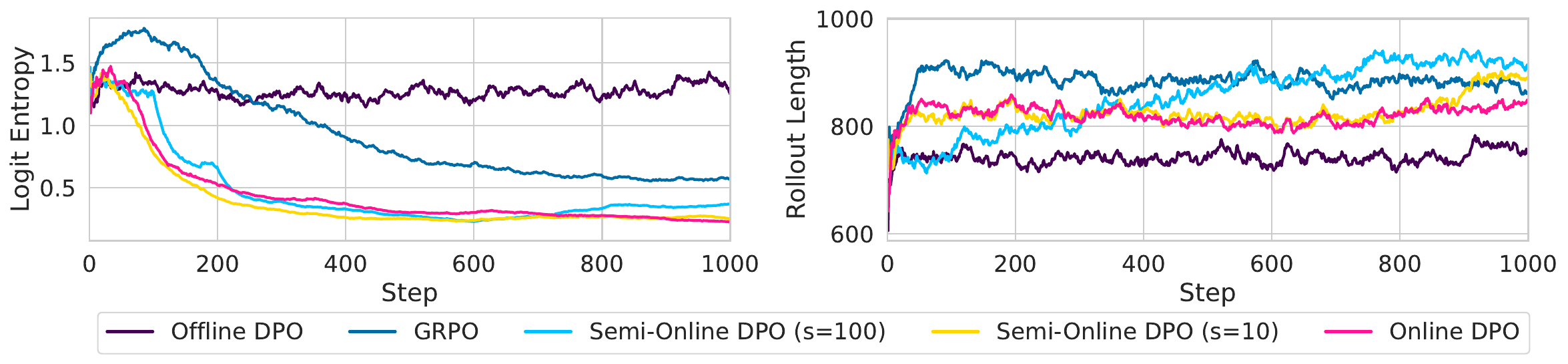}
    \caption{\textbf{Logit entropy collapse in iterative and online training on verifiable tasks.} Despite stable average length of rollouts during training (right), the average entropy of the next token distribution (left) decreases significantly during the training in all training regimes except the offline one.
    }
    \label{fig:entropy_collapse}
\end{figure}

\subsection{Additional Experiments and Observations}
\label{sec:additional_experiments}
\textbf{Response length\ } 
Although past work has found that both offline and online post-training methods tend to encourage longer answers \citep{park-etal-2024-disentangling,singhal_2024_long,guo2025deepseek}, we encounter both length increase and decrease in our training.
In the verifiable task, for example, we observe that disabling reference model sync and increasing training speed lead to greater risk of response length collapse and performance degradation (\autoref{fig:length_collapse}). We hypothesize that the bimodal distribution of response lengths (one peak with very short responses, and one with very long responses) is a major contributor to this collapse (\autoref{fig:response_lengths}).

On the other hand, we observe tendencies towards response length increase in the non-verifiable reward experiments. Since we are using an off-the-shelf LLM reward model, the model tends to hack its length bias to maximize rewards \citep{singhal_2024_long}. Therefore, the response lengths generally increase over time in the online or semi-online settings (\autoref{fig:logit_entropy_verifiable_vs_nonverifiable}, right). There are several methods to mitigate this: creating or finetuning a reward model for less length bias, incorporating a length penalty in the loss \citep{wu2024meta, park2024disentangling}, or selecting checkpoints by normalizing for length. For simplicity across all experiments, we choose the last option and find that this selection method generalizes well. 

\textbf{Entropy collapse and regularization\ }
We measure the entropy of the next token distribution averaged over all tokens sampled in rollouts in both DPO and GPRO experiments. \autoref{fig:entropy_collapse} shows substantial entropy collapse regardless of algorithm in the verifiable task, except for offline DPO.
It is possible that offline DPO training is also reducing entropy, but it is not detected here as the measurement is on the rollouts that are not generated from the current model.
Non-verifiable tasks, however, exhibit less collapse as training continues (\autoref{fig:logit_entropy_verifiable_vs_nonverifiable}, left). This may be due to both the task properties (\emph{i.e.}, gradual improvements, non-binary rewards) and the use of a model-based reward. 

We experiment with entropy regularization in the verifiable task to mitigate the entropy collapse in DPO. The average negative entropy of next token distribution is added to the training objective with a configurable coefficient. Empirical results in multiple levels of scale reveal that maintaining stable entropy throughout online training is a non-trivial task, demonstrated in \autoref{fig:entropy_reg}, and requires further investigation which we leave for future work.\

\textbf{Experiments with the loss\ }
Prior work reports benefits of adding an NLL loss over the chosen response $y_c$ to iterative DPO training \citep{pang2024iterative,hong-etal-2024-orpo}. We experiment with adding the NLL term to our online and semi-online DPO configurations in the verifiable task. We did not observe any benefits after adding the extra term (\autoref{fig:dpo_nll}). We explain this observation with the fact that chosen log probabilities do not substantially decrease during training in our case, which was one of the motivations for adding the NLL term in previous works.

While GRPO trains on all generated responses, DPO only utilizes a pair of responses.
In an attempt at improving utilization of the set of generated responses in DPO training, we propose to pair each of the correct responses $\mathcal{Y}_c$ with each the incorrect responses $\mathcal{Y}_r$ in the verifiable task, thus making a \textit{group} of preference pairs. We then average DPO losses off all pairs to compute the \textit{GroupDPO} loss:
\begin{align}
    \mathcal{L}_\text{GroupDPO} (x,\mathcal{Y}_c,\mathcal{Y}_r) = \frac{1}{|\mathcal{Y}_c|\cdot |\mathcal{Y}_r|}\sum_{y_c \in \mathcal{Y}_c} \sum_{y_r \in \mathcal{Y}_r} \mathcal{L}_\text{DPO}(x, y_c,y_r)
\end{align}
We experimented with GroupDPO in the verifiable task setup, as presented in~\autoref{fig:dpo_grpo_hybrids}, and did not observe substantial changes in the performance compared with the model using a single preference pair chosen randomly.

Both group DPO and GRPO can learn from all responses in an online manner, but using very different loss functions.
This begs the question whether these two losses can be combined.
We implemented this loss as $\mathcal{L}_\text{combined} = \mathcal{L}_\text{GroupDPO}(x,\mathcal{Y}_c,\mathcal{Y}_r) + \alpha \mathcal{L}_\text{GRPO}(x, \mathcal{Y}_c \cup \mathcal{Y}_r)$ and train on the verifiable task. We compare the results against semi-DPO and online group DPO in~\autoref{fig:dpo_grpo_hybrids} and find no substantial difference in reward or entropy.
All relevant hyper-parameters used in additional experiments are provided in Appendix \autoref{sec:extra_hparam}.

\section{Discussion}
\subsection{Hyperparameter selection}
Throughout our experiments, we made several observations about tuning hyperparameters for stable training.
In general, we observed frequent instabilities in DPO training that makes the learning suddenly diverge.
We found that increasing the Adam epsilon value reduces such collapses and improves the overall stability.
The reason for this might be that an epsilon value that is too small  forces Adam to make relatively constant updates regardless of the actual gradient value, which could lead to noisy updates when the loss surface is flatter and the gradient is near zero.
Increasing epsilon leads to slower convergence, but it can be compensated by increasing the learning rate or gradient clipping. 

We experimented with the GRPO loss that has the length normalization, but found it to be less stable.
Such normalization will decrease the gradients from longer sequences, which might lead to learning biased towards shorter sequences.
We observed a similar trend when adding a length-normalized NLL term to DPO training, where it can boost probabilities of shorter responses more.

\subsection{Training Efficiency}
One of the key advantages of DPO is its efficiency in requiring only a single pair of responses for each training step on a given prompt. In online DPO, this efficiency remains, with the caveat of having to sample responses and create the pairs at each step. This streamlined approach contrasts with GRPO, which necessitates an entire group of responses, typically more than two, for each prompt. While traditional DPO might be seen as sample-inefficient due to its practice of discarding some responses, the simplicity of needing just two responses per training step can be advantageous. This efficiency reduces GPU memory overhead in the training step, making DPO a more scalable option in compute-constrained settings. 

The semi-online configuration brings another advantage. Since the generator model does not need to be synchronized with the trainer during each $s$ step interval, all user prompts from that interval can be annotated with the model's responses asynchronously and in parallel. Speed-up benefits are bound to the technical implementation of these asynchronous annotations, and will scale up as we increase $s$. Such a feature is likely to be particularly attractive in large model post-training where inference is more computationally expensive.

\section{Related Work}

\textbf{Reinforcement learning for LLMs\ } The landmark InstructGPT paper \citep{ouyang2022training}
showed how reinforcement learning from Human feedback (RLHF) \citep{christiano2017deep,ziegler2019fine} can be applied to train instruction following LLMs.
This pipeline consisted of the previously standard use of 
Supervised Fine-Tuning (SFT) training on prompt–response pairs labeled by humans, followed by training a reward model (RM) on  human preference rankings over different responses, and finally RL using PPO \citep{schulman2017proximalpolicyoptimizationalgorithms} with the resulting RM.

\textbf{Offline vs iterative vs online training for LLMs\ }
Proposed in 2023, DPO \citep{rafailov2024direct} removes the need for a reward model and directly optimizes for preferred outputs using given pairwise comparisons. This method is offline, and does not depend on producing responses from the model during training. Due to this simplicity and good performance on some tasks, this approach was widely adopted in the community \citep{tunstall2023zephyr,mixtralblogpost2023}. However further analysis still revealed a gap in performance with online methods like PPO \citep{xu2024dpo,chen2024preference}.
Approaches to make DPO semi-online by iterative training, recollecting preferences pairs with the updated model every iteration, showed much stronger results \citep{xu2023some,xiong2023gibbs,chen2024self,yuan2024self} than standard offline DPO. Completely online variants of DPO were also proposed in \citet{qi2024online,guo2024direct}. \citet{xu-etal-2024-bpo} also investigated the tradeoffs between iterative and fully online DPO, finding that semi-online DPO could outperform on-policy DPO when the reference model was synced more frequently. However, their analyses were limited to non-verifiable tasks and relied upon stabilizing online training by setting $\pi_\text{ref}$ as an ensemble of multiple sets of LoRA \citep{hu2022lora} weights rather than simply setting $\pi_\text{ref}$ to an intermediate checkpoint. \cite{liu2023statistical} sample preference pairs from the estimated optimal policy, which is closer to online DPO, but is not fully online. \citet{xiong2023iterative} demonstrate that RLHF algorithms, including DPO, in general benefit from online exploration.

\textbf{Non-verifiable vs. verifiable tasks and reasoning models\ }
Much of the work in instruction following training has relied on reward models due to the challenging nature of verifying general tasks including open QA, chat, summarization, and creative writing. With the advent of optimizing reasoning models  there has been a renewed interest in verifiable rewards where the task has a known, easily verifiable answer, e.g. short deterministic answers in math problems \citep{hendrycks2021measuring,li2024numinamath}. 
\citet{pang2024iterative} showed that Iterative DPO with verifiable rewards could be applied to this setting to improve chain-of-thought reasoning.  
\citet{lambert2024t} showed that full online RL could be applied with verifiable rewards in a similar setting.
DeepSeek-R1 \citep{guo2025deepseek} applied GRPO \citep{shao2024deepseekmath} at scale to this setting, in addition to training on non-verifiable tasks, producing a powerful LLM that can think before answering.
\citet{wu2024thinking} applied iterative DPO for training such thinking LLMs using non-verifiable tasks.

\section{Conclusion}

We explored the effectiveness of various LLM finetuning RL methods across different training paradigms: offline, semi-online, and online, on both verifiable and non-verifiable tasks. Our findings indicate that while offline methods like DPO offer simplicity and efficiency, they often lag in performance compared to their semi-online and online counterparts. 
We find that fully online DPO and GRPO perform comparably while significantly outperforming offline DPO.
Semi-online DPO, which synchronizes the model less frequently than online methods, bridges this gap, nearing the performance levels of fully online methods while allowing the possibility of increased efficiency. 
The effectiveness of semi-online algorithms makes them viable for agentic applications, where models can continue updating while waiting for long streams of complex rollouts \citep{silver2025welcome}. This approach builds on asynchronous reinforcement learning principles, demonstrating how offline RL methods like DPO can be adapted to asynchronous settings, what we call semi-online learning.

Additionally, we demonstrate the effectiveness of combining both verifiable and non-verifiable tasks during training. While directly training on only verifiable or non-verifiable tasks yielded limited benefits or even performance decreases on the opposite transfer tasks, starting from a verifiable or non-verifiable trained checkpoint and finetuning on the opposite data type led to significant improvements. Finally, fine-tuning the seed model with a mix of both reward signals achieved improvements on both types of tasks compared to the base model, affirming the benefits of a mixed reward training approach. However, questions remain on the precise best recipe.

Our work provides an exploratory analysis of LLM post-training regimes from offline to offline learning, facilitating further investigation into optimal strategies, particularly around multi-task settings. Due to computational constraints, our experiments are limited to one type of seed LLM model, and future explorations may consider other variants. Future work may explore combining more reward types, including other verifiers, and/or other reward models.

\bibliographystyle{plainnat}
\bibliography{neurips_2025}

\begin{thebibliography}{55}
\providecommand{\natexlab}[1]{#1}
\providecommand{\url}[1]{\texttt{#1}}
\expandafter\ifx\csname urlstyle\endcsname\relax
  \providecommand{\doi}[1]{doi: #1}\else
  \providecommand{\doi}{doi: \begingroup \urlstyle{rm}\Url}\fi

\bibitem[Balioglu et~al.(2023)Balioglu, Gleize, Kozhevnikov, Kulikov, Tran, and Yao]{balioglu2023fairseq2}
Can Balioglu, Martin Gleize, Artyom Kozhevnikov, Ilia Kulikov, Tuan Tran, and Julien Yao.
\newblock fairseq2, 2023.
\newblock URL \url{http://github.com/facebookresearch/fairseq2}.

\bibitem[Chen et~al.(2024{\natexlab{a}})Chen, Malladi, Zhang, Chen, Zhang, Ranganath, and Cho]{chen2024preference}
Angelica Chen, Sadhika Malladi, Lily~H Zhang, Xinyi Chen, Qiuyi Zhang, Rajesh Ranganath, and Kyunghyun Cho.
\newblock Preference learning algorithms do not learn preference rankings.
\newblock In \emph{The Thirty-eighth Annual Conference on Neural Information Processing Systems}, 2024{\natexlab{a}}.
\newblock URL \url{https://openreview.net/forum?id=YkJ5BuEXdD}.

\bibitem[Chen et~al.(2024{\natexlab{b}})Chen, Deng, Yuan, Ji, and Gu]{chen2024self}
Zixiang Chen, Yihe Deng, Huizhuo Yuan, Kaixuan Ji, and Quanquan Gu.
\newblock Self-play fine-tuning converts weak language models to strong language models.
\newblock \emph{arXiv preprint arXiv:2401.01335}, 2024{\natexlab{b}}.

\bibitem[Christiano et~al.(2017)Christiano, Leike, Brown, Martic, Legg, and Amodei]{christiano2017deep}
Paul~F Christiano, Jan Leike, Tom Brown, Miljan Martic, Shane Legg, and Dario Amodei.
\newblock Deep reinforcement learning from human preferences.
\newblock \emph{Advances in neural information processing systems}, 30, 2017.

\bibitem[Dubey et~al.(2024)Dubey, Jauhri, Pandey, Kadian, Al-Dahle, Letman, Mathur, Schelten, Yang, Fan, et~al.]{dubey2024llama}
Abhimanyu Dubey, Abhinav Jauhri, Abhinav Pandey, Abhishek Kadian, Ahmad Al-Dahle, Aiesha Letman, Akhil Mathur, Alan Schelten, Amy Yang, Angela Fan, et~al.
\newblock The llama 3 herd of models.
\newblock \emph{arXiv preprint arXiv:2407.21783}, 2024.
\newblock Llama 3.1 Community License Agreement.

\bibitem[Dubois et~al.(2024)Dubois, Galambosi, Liang, and Hashimoto]{dubois2024length}
Yann Dubois, Bal{\'a}zs Galambosi, Percy Liang, and Tatsunori~B Hashimoto.
\newblock Length-controlled alpacaeval: A simple way to debias automatic evaluators.
\newblock \emph{arXiv preprint arXiv:2404.04475}, 2024.

\bibitem[Frick et~al.(2024{\natexlab{a}})Frick, Jin, Li, Ganesan, Zhang, Jiao, and Zhu]{Athene2024}
Evan Frick, Peter Jin, Tianle Li, Karthik Ganesan, Jian Zhang, Jiantao Jiao, and Banghua Zhu.
\newblock Athene-70b: Redefining the boundaries of post-training for open models, July 2024{\natexlab{a}}.
\newblock URL \url{https://nexusflow.ai/blogs/athene}.

\bibitem[Frick et~al.(2024{\natexlab{b}})Frick, Li, Chen, Chiang, Angelopoulos, Jiao, Zhu, Gonzalez, and Stoica]{frick2024evaluaterewardmodelsrlhf}
Evan Frick, Tianle Li, Connor Chen, Wei-Lin Chiang, Anastasios~N. Angelopoulos, Jiantao Jiao, Banghua Zhu, Joseph~E. Gonzalez, and Ion Stoica.
\newblock How to evaluate reward models for rlhf, 2024{\natexlab{b}}.
\newblock URL \url{https://arxiv.org/abs/2410.14872}.

\bibitem[Gao et~al.(2023)Gao, Schulman, and Hilton]{gao2023scaling}
Leo Gao, John Schulman, and Jacob Hilton.
\newblock Scaling laws for reward model overoptimization.
\newblock In \emph{International Conference on Machine Learning}, pages 10835--10866. PMLR, 2023.

\bibitem[Guo et~al.(2025)Guo, Yang, Zhang, Song, Zhang, Xu, Zhu, Ma, Wang, Bi, et~al.]{guo2025deepseek}
Daya Guo, Dejian Yang, Haowei Zhang, Junxiao Song, Ruoyu Zhang, Runxin Xu, Qihao Zhu, Shirong Ma, Peiyi Wang, Xiao Bi, et~al.
\newblock Deepseek-r1: Incentivizing reasoning capability in llms via reinforcement learning.
\newblock \emph{arXiv preprint arXiv:2501.12948}, 2025.

\bibitem[Guo et~al.(2024)Guo, Zhang, Liu, Liu, Khalman, Llinares, Rame, Mesnard, Zhao, Piot, et~al.]{guo2024direct}
Shangmin Guo, Biao Zhang, Tianlin Liu, Tianqi Liu, Misha Khalman, Felipe Llinares, Alexandre Rame, Thomas Mesnard, Yao Zhao, Bilal Piot, et~al.
\newblock Direct language model alignment from online ai feedback.
\newblock \emph{arXiv preprint arXiv:2402.04792}, 2024.

\bibitem[Hendrycks et~al.(2021)Hendrycks, Burns, Kadavath, Arora, Basart, Tang, Song, and Steinhardt]{hendrycks2021measuring}
Dan Hendrycks, Collin Burns, Saurav Kadavath, Akul Arora, Steven Basart, Eric Tang, Dawn Song, and Jacob Steinhardt.
\newblock Measuring mathematical problem solving with the math dataset.
\newblock \emph{arXiv preprint arXiv:2103.03874}, 2021.
\newblock MIT license.

\bibitem[Hong et~al.(2024)Hong, Lee, and Thorne]{hong-etal-2024-orpo}
Jiwoo Hong, Noah Lee, and James Thorne.
\newblock {ORPO}: Monolithic preference optimization without reference model.
\newblock In Yaser Al-Onaizan, Mohit Bansal, and Yun-Nung Chen, editors, \emph{Proceedings of the 2024 Conference on Empirical Methods in Natural Language Processing}, pages 11170--11189, Miami, Florida, USA, November 2024. Association for Computational Linguistics.
\newblock \doi{10.18653/v1/2024.emnlp-main.626}.
\newblock URL \url{https://aclanthology.org/2024.emnlp-main.626/}.

\bibitem[Hu et~al.(2022)Hu, yelong shen, Wallis, Allen-Zhu, Li, Wang, Wang, and Chen]{hu2022lora}
Edward~J Hu, yelong shen, Phillip Wallis, Zeyuan Allen-Zhu, Yuanzhi Li, Shean Wang, Lu~Wang, and Weizhu Chen.
\newblock Lo{RA}: Low-rank adaptation of large language models.
\newblock In \emph{International Conference on Learning Representations}, 2022.
\newblock URL \url{https://openreview.net/forum?id=nZeVKeeFYf9}.

\bibitem[Kwon et~al.(2023)Kwon, Li, Zhuang, Sheng, Zheng, Yu, Gonzalez, Zhang, and Stoica]{kwon2023efficient}
Woosuk Kwon, Zhuohan Li, Siyuan Zhuang, Ying Sheng, Lianmin Zheng, Cody~Hao Yu, Joseph~E. Gonzalez, Hao Zhang, and Ion Stoica.
\newblock Efficient memory management for large language model serving with pagedattention.
\newblock In \emph{Proceedings of the ACM SIGOPS 29th Symposium on Operating Systems Principles}, 2023.

\bibitem[Lambert et~al.(2024)Lambert, Morrison, Pyatkin, Huang, Ivison, Brahman, Miranda, Liu, Dziri, Lyu, et~al.]{lambert2024t}
Nathan Lambert, Jacob Morrison, Valentina Pyatkin, Shengyi Huang, Hamish Ivison, Faeze Brahman, Lester James~V Miranda, Alisa Liu, Nouha Dziri, Shane Lyu, et~al.
\newblock T$\backslash$" ulu 3: Pushing frontiers in open language model post-training.
\newblock \emph{arXiv preprint arXiv:2411.15124}, 2024.

\bibitem[Li et~al.(2024{\natexlab{a}})Li, Beeching, Tunstall, Lipkin, Soletskyi, Huang, Rasul, Yu, Jiang, Shen, et~al.]{li2024numinamath}
Jia Li, Edward Beeching, Lewis Tunstall, Ben Lipkin, Roman Soletskyi, Shengyi Huang, Kashif Rasul, Longhui Yu, Albert~Q Jiang, Ziju Shen, et~al.
\newblock Numinamath: The largest public dataset in ai4maths with 860k pairs of competition math problems and solutions.
\newblock \emph{Hugging Face repository}, 13:\penalty0 9, 2024{\natexlab{a}}.

\bibitem[Li et~al.(2024{\natexlab{b}})Li, Chiang, Frick, Dunlap, Wu, Zhu, Gonzalez, and Stoica]{arenahard2024}
Tianle Li, Wei-Lin Chiang, Evan Frick, Lisa Dunlap, Tianhao Wu, Banghua Zhu, Joseph~E Gonzalez, and Ion Stoica.
\newblock From live data to high-quality benchmarks: The arena-hard pipeline, April 2024{\natexlab{b}}.
\newblock URL \url{https://lmsys.org/blog/2024-04-19-arena-hard/}.
\newblock Apache-2.0 license.

\bibitem[Li et~al.(2024{\natexlab{c}})Li, Chiang, Frick, Dunlap, Wu, Zhu, Gonzalez, and Stoica]{li2024crowdsourced}
Tianle Li, Wei-Lin Chiang, Evan Frick, Lisa Dunlap, Tianhao Wu, Banghua Zhu, Joseph~E Gonzalez, and Ion Stoica.
\newblock From crowdsourced data to high-quality benchmarks: Arena-hard and benchbuilder pipeline.
\newblock \emph{arXiv preprint arXiv:2406.11939}, 2024{\natexlab{c}}.

\bibitem[Li et~al.(2023{\natexlab{a}})Li, Yu, Zhou, Schick, Levy, Zettlemoyer, Weston, and Lewis]{li2023self}
Xian Li, Ping Yu, Chunting Zhou, Timo Schick, Omer Levy, Luke Zettlemoyer, Jason Weston, and Mike Lewis.
\newblock Self-alignment with instruction backtranslation.
\newblock \emph{arXiv preprint arXiv:2308.06259}, 2023{\natexlab{a}}.

\bibitem[Li et~al.(2023{\natexlab{b}})Li, Zhang, Dubois, Taori, Gulrajani, Guestrin, Liang, and Hashimoto]{alpaca_eval}
Xuechen Li, Tianyi Zhang, Yann Dubois, Rohan Taori, Ishaan Gulrajani, Carlos Guestrin, Percy Liang, and Tatsunori~B. Hashimoto.
\newblock Alpacaeval: An automatic evaluator of instruction-following models.
\newblock \url{https://github.com/tatsu-lab/alpaca_eval}, 5 2023{\natexlab{b}}.
\newblock Apache-2.0 license.

\bibitem[Lightman et~al.(2023)Lightman, Kosaraju, Burda, Edwards, Baker, Lee, Leike, Schulman, Sutskever, and Cobbe]{lightman2023lets}
Hunter Lightman, Vineet Kosaraju, Yura Burda, Harri Edwards, Bowen Baker, Teddy Lee, Jan Leike, John Schulman, Ilya Sutskever, and Karl Cobbe.
\newblock Let's verify step by step.
\newblock \emph{arXiv preprint arXiv:2305.20050}, 2023.

\bibitem[Liu et~al.(2023)Liu, Zhao, Joshi, Khalman, Saleh, Liu, and Liu]{liu2023statistical}
Tianqi Liu, Yao Zhao, Rishabh Joshi, Misha Khalman, Mohammad Saleh, Peter~J Liu, and Jialu Liu.
\newblock Statistical rejection sampling improves preference optimization.
\newblock \emph{arXiv preprint arXiv:2309.06657}, 2023.

\bibitem[Liu et~al.(2025)Liu, Chen, Li, Qi, Pang, Du, Lee, and Lin]{liu2025understanding}
Zichen Liu, Changyu Chen, Wenjun Li, Penghui Qi, Tianyu Pang, Chao Du, Wee~Sun Lee, and Min Lin.
\newblock Understanding r1-zero-like training: A critical perspective, 2025.

\bibitem[Mistral AI team, 2023()]{mixtralblogpost2023}
Mistral AI team, 2023.
\newblock Mixtral of experts: A high quality sparse mixture-of-experts.
\newblock \url{https://mistral.ai/news/mixtral-of-experts/}, 2023.
\newblock Accessed: Dec 12th 2023.

\bibitem[Moritz et~al.(2018)Moritz, Nishihara, Wang, Tumanov, Liaw, Liang, Elibol, Yang, Paul, Jordan, et~al.]{moritz2018ray}
Philipp Moritz, Robert Nishihara, Stephanie Wang, Alexey Tumanov, Richard Liaw, Eric Liang, Melih Elibol, Zongheng Yang, William Paul, Michael~I Jordan, et~al.
\newblock Ray: A distributed framework for emerging $\{$AI$\}$ applications.
\newblock In \emph{13th USENIX symposium on operating systems design and implementation (OSDI 18)}, pages 561--577, 2018.

\bibitem[Ouyang et~al.(2022)Ouyang, Wu, Jiang, Almeida, Wainwright, Mishkin, Zhang, Agarwal, Slama, Ray, et~al.]{ouyang2022training}
Long Ouyang, Jeffrey Wu, Xu~Jiang, Diogo Almeida, Carroll Wainwright, Pamela Mishkin, Chong Zhang, Sandhini Agarwal, Katarina Slama, Alex Ray, et~al.
\newblock Training language models to follow instructions with human feedback.
\newblock \emph{Advances in neural information processing systems}, 35:\penalty0 27730--27744, 2022.

\bibitem[Pace et~al.(2024)Pace, Mallinson, Malmi, Krause, and Severyn]{pace2024west}
Aliz{\'e}e Pace, Jonathan Mallinson, Eric Malmi, Sebastian Krause, and Aliaksei Severyn.
\newblock West-of-n: Synthetic preference generation for improved reward modeling.
\newblock \emph{arXiv preprint arXiv:2401.12086}, 2024.

\bibitem[Pang et~al.(2024)Pang, Yuan, He, Cho, Sukhbaatar, and Weston]{pang2024iterative}
Richard~Yuanzhe Pang, Weizhe Yuan, He~He, Kyunghyun Cho, Sainbayar Sukhbaatar, and Jason Weston.
\newblock Iterative reasoning preference optimization.
\newblock \emph{Advances in Neural Information Processing Systems}, 37:\penalty0 116617--116637, 2024.

\bibitem[Park et~al.(2024{\natexlab{a}})Park, Rafailov, Ermon, and Finn]{park-etal-2024-disentangling}
Ryan Park, Rafael Rafailov, Stefano Ermon, and Chelsea Finn.
\newblock Disentangling length from quality in direct preference optimization.
\newblock In Lun-Wei Ku, Andre Martins, and Vivek Srikumar, editors, \emph{Findings of the Association for Computational Linguistics: ACL 2024}, pages 4998--5017, Bangkok, Thailand, August 2024{\natexlab{a}}. Association for Computational Linguistics.
\newblock \doi{10.18653/v1/2024.findings-acl.297}.
\newblock URL \url{https://aclanthology.org/2024.findings-acl.297/}.

\bibitem[Park et~al.(2024{\natexlab{b}})Park, Rafailov, Ermon, and Finn]{park2024disentangling}
Ryan Park, Rafael Rafailov, Stefano Ermon, and Chelsea Finn.
\newblock Disentangling length from quality in direct preference optimization.
\newblock \emph{arXiv preprint arXiv:2403.19159}, 2024{\natexlab{b}}.

\bibitem[Qi et~al.(2024)Qi, Li, Li, Gao, Zhang, and Zhou]{qi2024online}
Biqing Qi, Pengfei Li, Fangyuan Li, Junqi Gao, Kaiyan Zhang, and Bowen Zhou.
\newblock Online dpo: Online direct preference optimization with fast-slow chasing, 2024.

\bibitem[Rafailov et~al.(2023)Rafailov, Sharma, Mitchell, Manning, Ermon, and Finn]{rafailov2023direct}
Rafael Rafailov, Archit Sharma, Eric Mitchell, Christopher~D Manning, Stefano Ermon, and Chelsea Finn.
\newblock Direct preference optimization: Your language model is secretly a reward model.
\newblock \emph{Advances in Neural Information Processing Systems}, 36:\penalty0 53728--53741, 2023.

\bibitem[Rafailov et~al.(2024)Rafailov, Sharma, Mitchell, Manning, Ermon, and Finn]{rafailov2024direct}
Rafael Rafailov, Archit Sharma, Eric Mitchell, Christopher~D Manning, Stefano Ermon, and Chelsea Finn.
\newblock Direct preference optimization: Your language model is secretly a reward model.
\newblock \emph{Advances in Neural Information Processing Systems}, 36, 2024.

\bibitem[Schulman et~al.(2015)Schulman, Levine, Abbeel, Jordan, and Moritz]{schulman2015trust}
John Schulman, Sergey Levine, Pieter Abbeel, Michael Jordan, and Philipp Moritz.
\newblock Trust region policy optimization.
\newblock In \emph{International conference on machine learning}, pages 1889--1897. PMLR, 2015.

\bibitem[Schulman et~al.(2017{\natexlab{a}})Schulman, Wolski, Dhariwal, Radford, and Klimov]{schulman2017proximal}
John Schulman, Filip Wolski, Prafulla Dhariwal, Alec Radford, and Oleg Klimov.
\newblock Proximal policy optimization algorithms.
\newblock \emph{arXiv preprint arXiv:1707.06347}, 2017{\natexlab{a}}.

\bibitem[Schulman et~al.(2017{\natexlab{b}})Schulman, Wolski, Dhariwal, Radford, and Klimov]{schulman2017proximalpolicyoptimizationalgorithms}
John Schulman, Filip Wolski, Prafulla Dhariwal, Alec Radford, and Oleg Klimov.
\newblock Proximal policy optimization algorithms, 2017{\natexlab{b}}.
\newblock URL \url{https://arxiv.org/abs/1707.06347}.

\bibitem[Shao et~al.(2024)Shao, Wang, Zhu, Xu, Song, Bi, Zhang, Zhang, Li, Wu, et~al.]{shao2024deepseekmath}
Zhihong Shao, Peiyi Wang, Qihao Zhu, Runxin Xu, Junxiao Song, Xiao Bi, Haowei Zhang, Mingchuan Zhang, YK~Li, Y~Wu, et~al.
\newblock Deepseekmath: Pushing the limits of mathematical reasoning in open language models.
\newblock \emph{arXiv preprint arXiv:2402.03300}, 2024.

\bibitem[Silver and Sutton(2025)]{silver2025welcome}
David Silver and Richard~S Sutton.
\newblock Welcome to the era of experience.
\newblock \emph{Google AI}, 1, 2025.

\bibitem[Singhal et~al.(2024)Singhal, Goyal, Xu, and Durrett]{singhal_2024_long}
Prasann Singhal, Tanya Goyal, Jiacheng Xu, and Greg Durrett.
\newblock A long way to go: Investigating length correlations in {RLHF}.
\newblock In \emph{First Conference on Language Modeling}, 2024.
\newblock URL \url{https://openreview.net/forum?id=G8LaO1P0xv}.

\bibitem[Touvron et~al.(2023{\natexlab{a}})Touvron, Lavril, Izacard, Martinet, Lachaux, Lacroix, Rozi{\`e}re, Goyal, Hambro, Azhar, et~al.]{touvron2023llama1}
Hugo Touvron, Thibaut Lavril, Gautier Izacard, Xavier Martinet, Marie-Anne Lachaux, Timoth{\'e}e Lacroix, Baptiste Rozi{\`e}re, Naman Goyal, Eric Hambro, Faisal Azhar, et~al.
\newblock Llama: Open and efficient foundation language models.
\newblock \emph{arXiv preprint arXiv:2302.13971}, 2023{\natexlab{a}}.

\bibitem[Touvron et~al.(2023{\natexlab{b}})Touvron, Martin, Stone, Albert, Almahairi, Babaei, Bashlykov, Batra, Bhargava, Bhosale, et~al.]{touvron2023llama}
Hugo Touvron, Louis Martin, Kevin Stone, Peter Albert, Amjad Almahairi, Yasmine Babaei, Nikolay Bashlykov, Soumya Batra, Prajjwal Bhargava, Shruti Bhosale, et~al.
\newblock Llama 2: Open foundation and fine-tuned chat models.
\newblock \emph{arXiv preprint arXiv:2307.09288}, 2023{\natexlab{b}}.

\bibitem[Tunstall et~al.(2023)Tunstall, Beeching, Lambert, Rajani, Rasul, Belkada, Huang, von Werra, Fourrier, Habib, et~al.]{tunstall2023zephyr}
Lewis Tunstall, Edward Beeching, Nathan Lambert, Nazneen Rajani, Kashif Rasul, Younes Belkada, Shengyi Huang, Leandro von Werra, Cl{\'e}mentine Fourrier, Nathan Habib, et~al.
\newblock Zephyr: Direct distillation of lm alignment.
\newblock \emph{arXiv preprint arXiv:2310.16944}, 2023.

\bibitem[Wu et~al.(2024{\natexlab{a}})Wu, Lan, Yuan, Jiao, Weston, and Sukhbaatar]{wu2024thinking}
Tianhao Wu, Janice Lan, Weizhe Yuan, Jiantao Jiao, Jason Weston, and Sainbayar Sukhbaatar.
\newblock Thinking llms: General instruction following with thought generation.
\newblock \emph{arXiv preprint arXiv:2410.10630}, 2024{\natexlab{a}}.

\bibitem[Wu et~al.(2024{\natexlab{b}})Wu, Yuan, Golovneva, Xu, Tian, Jiao, Weston, and Sukhbaatar]{wu2024meta}
Tianhao Wu, Weizhe Yuan, Olga Golovneva, Jing Xu, Yuandong Tian, Jiantao Jiao, Jason Weston, and Sainbayar Sukhbaatar.
\newblock Meta-rewarding language models: Self-improving alignment with llm-as-a-meta-judge.
\newblock \emph{arXiv preprint arXiv:2407.19594}, 2024{\natexlab{b}}.

\bibitem[Xiong et~al.(2023{\natexlab{a}})Xiong, Dong, Ye, Wang, Zhong, Ji, Jiang, and Zhang]{xiong2023iterative}
Wei Xiong, Hanze Dong, Chenlu Ye, Ziqi Wang, Han Zhong, Heng Ji, Nan Jiang, and Tong Zhang.
\newblock Iterative preference learning from human feedback: Bridging theory and practice for rlhf under kl-constraint.
\newblock \emph{arXiv preprint arXiv:2312.11456}, 2023{\natexlab{a}}.

\bibitem[Xiong et~al.(2023{\natexlab{b}})Xiong, Dong, Ye, Zhong, Jiang, and Zhang]{xiong2023gibbs}
Wei Xiong, Hanze Dong, Chenlu Ye, Han Zhong, Nan Jiang, and Tong Zhang.
\newblock Gibbs sampling from human feedback: A provable kl-constrained framework for rlhf.
\newblock \emph{CoRR}, 2023{\natexlab{b}}.

\bibitem[Xu et~al.(2023{\natexlab{a}})Xu, Sun, Zheng, Geng, Zhao, Feng, Tao, and Jiang]{xu2023wizardlm}
Can Xu, Qingfeng Sun, Kai Zheng, Xiubo Geng, Pu~Zhao, Jiazhan Feng, Chongyang Tao, and Daxin Jiang.
\newblock Wizardlm: Empowering large language models to follow complex instructions.
\newblock \emph{arXiv preprint arXiv:2304.12244}, 2023{\natexlab{a}}.

\bibitem[Xu et~al.(2023{\natexlab{b}})Xu, Lee, Sukhbaatar, and Weston]{xu2023some}
Jing Xu, Andrew Lee, Sainbayar Sukhbaatar, and Jason Weston.
\newblock Some things are more cringe than others: Preference optimization with the pairwise cringe loss.
\newblock \emph{arXiv preprint arXiv:2312.16682}, 2023{\natexlab{b}}.

\bibitem[Xu et~al.(2024{\natexlab{a}})Xu, Fu, Gao, Ye, Liu, Mei, Wang, Yu, and Wu]{xu2024dpo}
Shusheng Xu, Wei Fu, Jiaxuan Gao, Wenjie Ye, Weilin Liu, Zhiyu Mei, Guangju Wang, Chao Yu, and Yi~Wu.
\newblock Is dpo superior to ppo for llm alignment? a comprehensive study.
\newblock \emph{arXiv preprint arXiv:2404.10719}, 2024{\natexlab{a}}.

\bibitem[Xu et~al.(2024{\natexlab{b}})Xu, Li, Wang, and Li]{xu-etal-2024-bpo}
Wenda Xu, Jiachen Li, William~Yang Wang, and Lei Li.
\newblock {BPO}: Staying close to the behavior {LLM} creates better online {LLM} alignment.
\newblock In Yaser Al-Onaizan, Mohit Bansal, and Yun-Nung Chen, editors, \emph{Proceedings of the 2024 Conference on Empirical Methods in Natural Language Processing}, pages 11125--11139, Miami, Florida, USA, November 2024{\natexlab{b}}. Association for Computational Linguistics.
\newblock \doi{10.18653/v1/2024.emnlp-main.623}.
\newblock URL \url{https://aclanthology.org/2024.emnlp-main.623/}.

\bibitem[Yuan et~al.(2024)Yuan, Pang, Cho, Sukhbaatar, Xu, and Weston]{yuan2024self}
Weizhe Yuan, Richard~Yuanzhe Pang, Kyunghyun Cho, Sainbayar Sukhbaatar, Jing Xu, and Jason Weston.
\newblock Self-rewarding language models.
\newblock \emph{arXiv preprint arXiv:2401.10020}, 2024.

\bibitem[Zhao et~al.(2024)Zhao, Ren, Hessel, Cardie, Choi, and Deng]{zhao2024wildchat}
Wenting Zhao, Xiang Ren, Jack Hessel, Claire Cardie, Yejin Choi, and Yuntian Deng.
\newblock Wildchat: 1m chatgpt interaction logs in the wild.
\newblock \emph{arXiv preprint arXiv:2405.01470}, 2024.
\newblock Open Data Commons License Attribution family License.

\bibitem[Zhou et~al.(2023)Zhou, Liu, Xu, Iyer, Sun, Mao, Ma, Efrat, Yu, Yu, et~al.]{zhou2023lima}
Chunting Zhou, Pengfei Liu, Puxin Xu, Srinivasan Iyer, Jiao Sun, Yuning Mao, Xuezhe Ma, Avia Efrat, Ping Yu, Lili Yu, et~al.
\newblock Lima: Less is more for alignment.
\newblock \emph{Advances in Neural Information Processing Systems}, 36:\penalty0 55006--55021, 2023.

\bibitem[Ziegler et~al.(2019)Ziegler, Stiennon, Wu, Brown, Radford, Amodei, Christiano, and Irving]{ziegler2019fine}
Daniel~M Ziegler, Nisan Stiennon, Jeffrey Wu, Tom~B Brown, Alec Radford, Dario Amodei, Paul Christiano, and Geoffrey Irving.
\newblock Fine-tuning language models from human preferences.
\newblock \emph{arXiv preprint arXiv:1909.08593}, 2019.

\end{thebibliography}

\appendix
\clearpage
\section{Extended Background}\label{sec:extended_background}
\subsection{LLM Alignment Algorithms}\label{sec:extended_background_alignment}
\textbf{Direct Preference Optimization (DPO)} DPO \citep{rafailov2024direct} starts with optimizing the expected sequence-level reward, $r(y)$, with an additional KL term
\begin{align}
    \mathcal{O} = \mathbb{E}_{y\sim \pi} \left[ r(y) \right] - \beta \text{KL} \left[ \pi(y|x) || \piref(y|x)\right]
\end{align}
where $\piref$ is a reference model (the seed model by default). This objective can be converted into a single KL term (see \citet{rafailov2023direct} for proof):
\begin{align}
    \mathcal{O} = -\text{KL} \left[ \pi(y|x) || \pi^*(y|x)\right], \quad \text{where}\  \pi^*(y|x) = \piref(y|x) e^{r(y)/\beta}.
\end{align}
This shows that the optimal policy for rewards $r(y|x)$ is $\pi^*(y|x)$. Now let us write down the reward that corresponds to the current policy: $r'(y|x) = \beta \log \pi(y|x)/\piref(y|x)$.
DPO is designed to learn from preference labels $y_c \succ y_r$ where response $y_c$ is deemed better than $y_r$ for a given prompt $x$. 
This relation is then converted into rewards by the Bradley-Terry Model 
\begin{align}
p(y_c \succ y_r|x) &= \sigma (r'(y_c|x) - r'(y_r|x)) = \sigma \left( \beta \log \frac{\pi(y_c|x)}{\piref(y_c|x)} - \beta \log \frac{\pi(y_r|x)}{\piref(y_r|x)} \right) .
\end{align}
One can now optimize this using a cross-entropy loss, which gives us the DPO loss
\begin{align}
\mathcal{L}_\text{DPO}= -\log \sigma \left( \beta \log \frac{\pi(y_c|x)}{\piref(y_c|x)} - \beta \log \frac{\pi(y_r|x)}{\piref(y_r|x)} \right) .
\end{align}

\paragraph{PPO} PPO \citep{schulman2015trust} is an on-policy policy-gradient method that optimizes 
\begin{equation}
\mathcal{L}_\text{PPO} =- \mathbb{E}_{y \sim\piold} \left[ \sum_t \min \left\{ \frac{\pic(y_t|x,y_{<t})}{\piold(y_t|x,y_{<t})} A_t, \text{clip}_\epsilon\left( \frac{\pic(y_t|x,y_{<t})}{\piold(y_t|x,y_{<t})} \right) A_t \right\}  \right]  . 
\end{equation}

\section{Training Details}

\subsection{Online recipe technical details}
\label{sec:design_details}

The primary challenge in online training is to enable efficient inference using the latest policy model parameters and optionally the LLM-based reward model. 
In the framework pipeline, fairseq2's trainer runs as a standard \textit{single program multiple data} (SPMD) run, while generator, reference and reward models run as Ray actors \citep{moritz2018ray} on a standalone Ray cluster. This design allows us to plug in multiple reward models without sacrificing memory capacity of the trainer. Model weight synchronization is done directly between GPU devices using NCCL, and generation communication is done via Ray.

Our training pipeline runs as follows. The process begins with generating policy responses. These responses are then sent to a rewarding unit to compute rewards using either a rule-based system for verifiable tasks or a (LLM-based) reward model. Once the rewards are calculated, a preference or reward batch is composed using the corresponding preference tuning algorithm. This batch is then sent to the preference optimization (DPO) or RL (GRPO) unit to complete the training step.

\subsection{Additional experiments hyperparameters}
\label{sec:extra_hparam}

In the experiments with adding an NLL loss term we used NLL scale $1.0$. In the experiment with combining GroupDPO and GRPO objectives we tried to scale GRPO loss using scales from a set $\{0.01, 0.001\}$. In the experiments with entropy regularization we tried regularizer scales from a set $\{0.0001, 0.0002, 0.0003, 0.0005\}$.

\begin{table}[htbp]
  \centering
  \caption{Hyperparameter Settings for Different Tasks. }
  \label{tab:training_hyperparameters}
  \resizebox{1.0\linewidth}{!}{%
  \begin{tabular}{p{4.9cm}llccccccc}
    \toprule
    \textbf{Task Type} & \textbf{Task }& \textbf{\shortstack{KL\\$\beta$}} & \textbf{\shortstack{Learning \\ Rate}} & \textbf{\shortstack{Adam\\$\epsilon$}} & \textbf{\shortstack{Grad\\Clip}} & \textbf{\shortstack{Ref model\\Sync}} & \textbf{\shortstack{Max\\Len.}} & \textbf{\shortstack{Batch\\Sz.}} \\
    \midrule
    \multirow{4}{*}{Verifiable}
    & Offline DPO & 0.1 & 1e-6 & 1e-4 & 1.0 & No & 2048 & 64 \\
    & Semi-Online DPO & 0.1 & 1e-6 & 1e-4 & 1.0 & Yes & 2048 & 64 \\
    & Online DPO & 0.1 & 1e-6 & 1e-4 & 1.0 & Yes & 2048 & 64 \\
    & GRPO & 0.001 & 1e-6 & 1e-4 & 1.0 & No & 2048 & 64 \\
    \midrule
    \multirow{4}{*}{Non-Verifiable}
    & Offline DPO & 0.01 & 1e-6 & 1e-8 & 0.1 & No & 1024 & 32 \\
    & Semi-Online DPO & 0.01 & 1e-6 & 1e-8 & 0.1 & No & 1024 & 32\\
    & Online DPO & 0.01 & 1e-6 & 1e-8 & 0.1 & No & 1024 & 32\\
    & GRPO & 0.001 & 1e-6 & 1e-8 & 0.1 & No & 1024 & 32 \\
    \midrule
    Verifiable+Non-Verifiable \\ (NM-Ckpt, WC only) & \multirow{2}{*}[3.9ex]{Online DPO} & \multirow{2}{*}[3.9ex]{0.01} & \multirow{2}{*}[3.9ex]{1e-6} & \multirow{2}{*}[3.9ex]{1e-8} & \multirow{2}{*}[3.9ex]{0.1} & \multirow{2}{*}[3.9ex]{No} & \multirow{2}{*}[3.9ex]{2048} & \multirow{2}{*}[3.9ex]{32} \\[0.3em]
    
    Verifiable+Non-Verifiable \\ (WC-Ckpt, NM only) & \multirow{2}{*}[3.9ex]{Online DPO} & \multirow{2}{*}[3.9ex]{0.1} & \multirow{2}{*}[3.9ex]{1e-6} & \multirow{2}{*}[3.9ex]{1e-4} & \multirow{2}{*}[3.9ex]{1.0} & \multirow{2}{*}[3.9ex]{Yes} & \multirow{2}{*}[3.9ex]{2048} & \multirow{2}{*}[3.9ex]{32} \\
    [0.3em]
    
    Verifiable+Non-Verifiable \\ (Llama-3.1-8B-Instr., NM+WC)  & \multirow{2}{*}[3.9ex]{Online DPO}  & \multirow{2}{*}[3.9ex]{0.01} & \multirow{2}{*}[3.9ex]{1e-6} & \multirow{2}{*}[3.9ex]{1e-5} & \multirow{2}{*}[3.9ex]{0.1} & \multirow{2}{*}[3.9ex]{No} & \multirow{2}{*}[3.9ex]{2048} & \multirow{2}{*}[3.9ex]{32} \\
    \bottomrule
  \end{tabular}
  }
\end{table}

\begin{figure*}[t]
    \centering
    \begin{prompt}{Verifiable task prompt}
    \texttt{<|start\_header\_id|>user<|end\_header\_id|>}

\texttt{Given the following problem, reason and give a final answer to the problem.}
\texttt{Problem: \{PROBLEM\}}

\texttt{Your response should end with `The final answer is \$\\boxed\{[answer]\}\$. I hope it is correct.' where [answer] is the response to the problem.<|eot\_id|><|start\_header\_id|>assistant<|end\_header\_id|>}
\end{prompt}

\vspace{-1em}
    \caption{LLM prompt used for verifiable task.
    \label{fig:verifiable_prompt}
    }
\end{figure*}

\begin{figure*}[t]
    \centering
    \begin{prompt}{Non-verifiable task prompt}
    \texttt{<|start\_header\_id|>user<|end\_header\_id|>}

\texttt{\{WILDCHAT INSTRUCTION\} <|eot\_id|><|start\_header\_id|>assistant<|end\_header\_id|>}
\end{prompt}

\vspace{-1em}
    \caption{LLM prompt used for the non-verifiable task.
    \label{fig:non-verifiable_prompt}
    }
\end{figure*}

\section{Additional Experimental Results}

\begin{table}[]
  \centering
\caption{Verifiable task results showing acc (std error). Sampling temperature is set to 0.6, top-p is set to 0.9. Standard error has been computed over $N=50$ random seeds.}
\label{tab:grpo_n_ablation}
 \resizebox{0.7\textwidth}{!}{
  \begin{tabular}{@{}l c c c@{}}
    \toprule
    \textbf{Training method}                 & \textbf{Math500}         & \textbf{NuminaMath}    & \textbf{AMC23}           \\
    \midrule
    Seed (\texttt{Llama-3.1-8B-Instruct}) & 47.4 (1.6)   & 33.9 (0.6)   & 23.7 (5.2)   \\
    \midrule
    GRPO n=4              & 55.7 (1.4)   & 37.7 (0.5)   & 30.6 (5.2)   \\
    GRPO n=8              & 58.1 (1.3)   & 38.8 (0.5)   & 33.6 (5.1)   \\
    GRPO n=12             & 57.6 (1.2)   & 38.4 (0.6)   & 32.2 (5.9)   \\
    \bottomrule
  \end{tabular}
  }
  \vspace{1pt}
\end{table}

\begin{figure}[ht!]
    \centering
    \includegraphics[width=\linewidth]{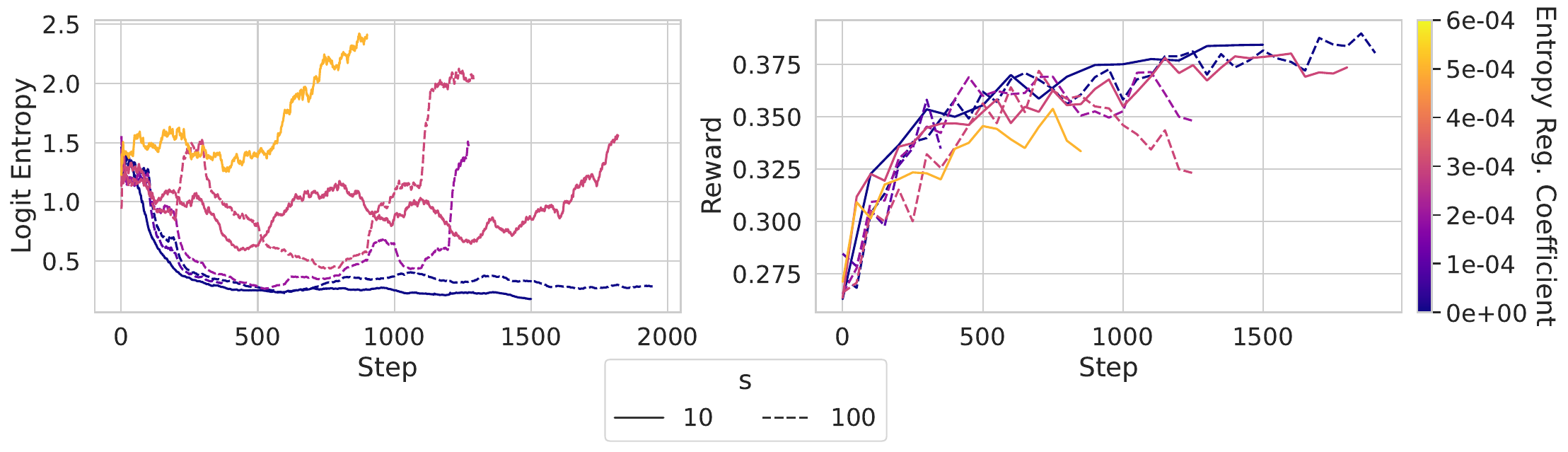}
    \caption{Logit entropy of rollouts and validation rewards of semi-online DPO with (coefficient $>0$) and without (coefficient $=0$) entropy regularization. Line color indicates strength of the regularization and line style indicates sync intervals.}
    \label{fig:entropy_reg}
\end{figure}

\begin{figure}[ht!]
    \centering
    \includegraphics[width=\linewidth]{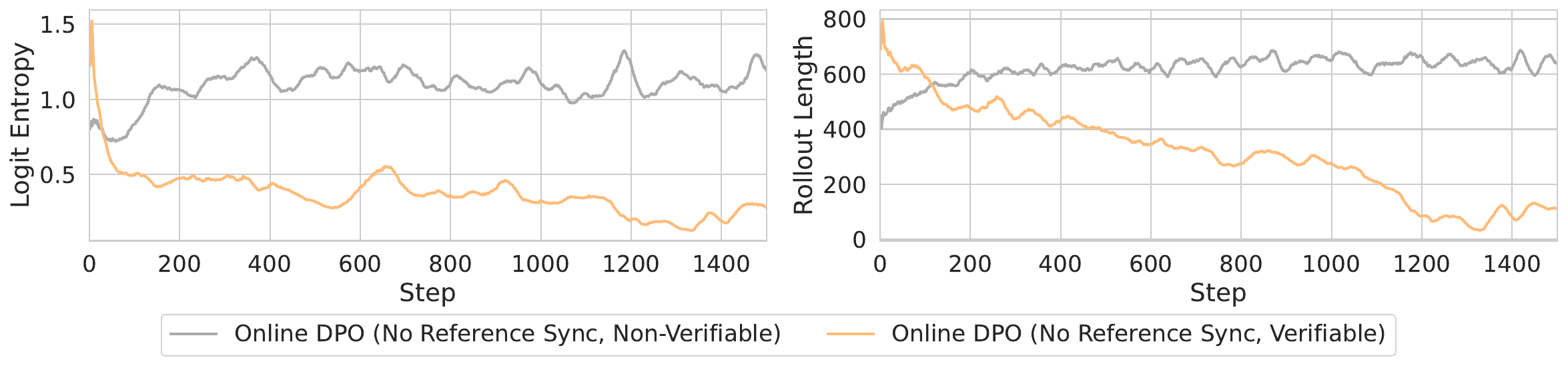}
    \caption{The logit entropy of online DPO trained without reference model synchronization is more likely to collapse when trained on verifiable tasks than on non-verifiable tasks.}
    \label{fig:logit_entropy_verifiable_vs_nonverifiable}
\end{figure}

\begin{figure}[ht!]
    \centering
    \includegraphics[width=\linewidth]{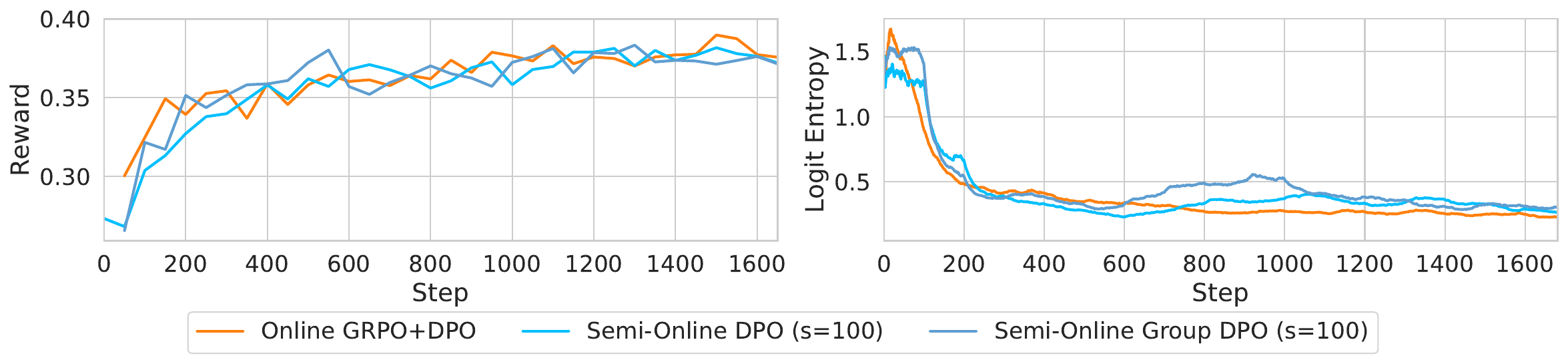}
    \caption{Validation reward and logit entropy of Group DPO, and combining GRPO and DPO compared against semi-online DPO.}
    \label{fig:dpo_grpo_hybrids}
\end{figure}

\begin{figure}[ht!]
    \centering
    \begin{subfigure}[t]{0.47\textwidth}
        \includegraphics[width=\textwidth]{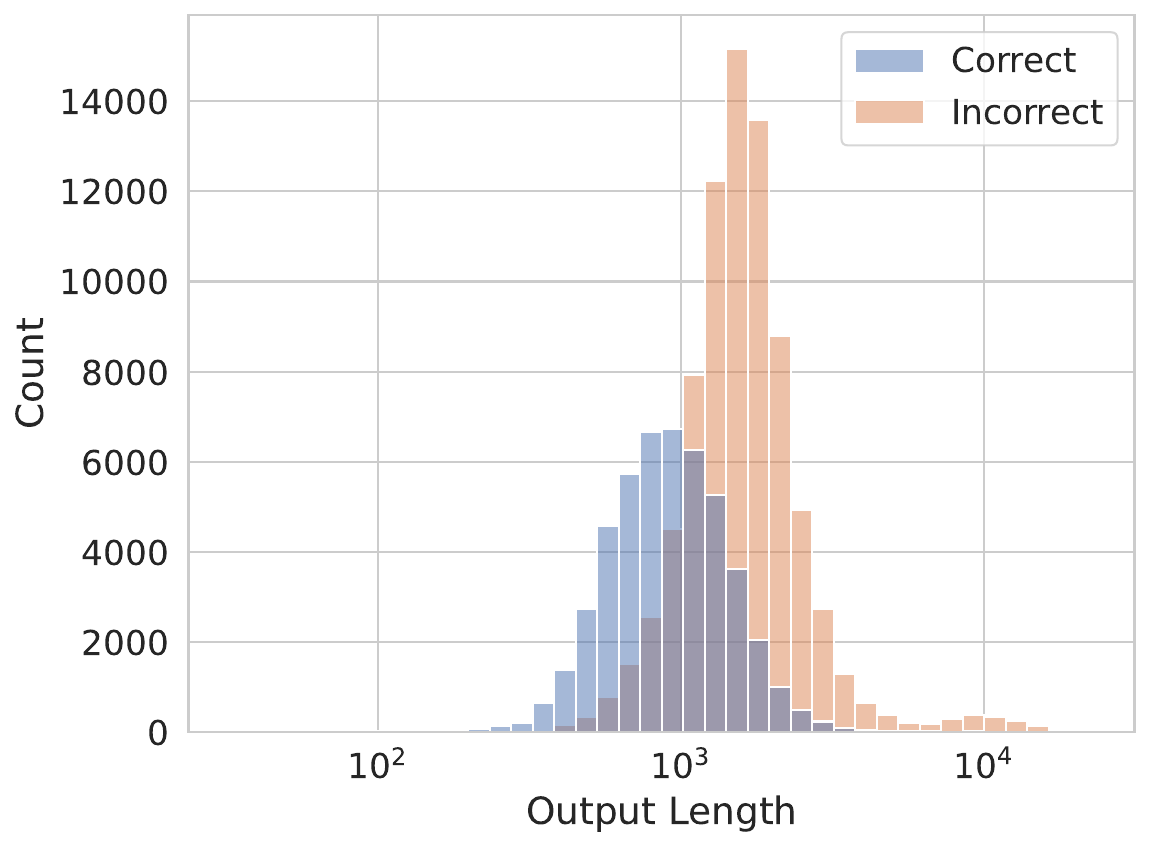}
        \caption{Output lengths of the first checkpoint of the collapsed online DPO run (with no reference model sync) on all math benchmarks.}
    \end{subfigure}
    \hfill
    \begin{subfigure}[t]{0.47\textwidth}
        \includegraphics[width=\textwidth]{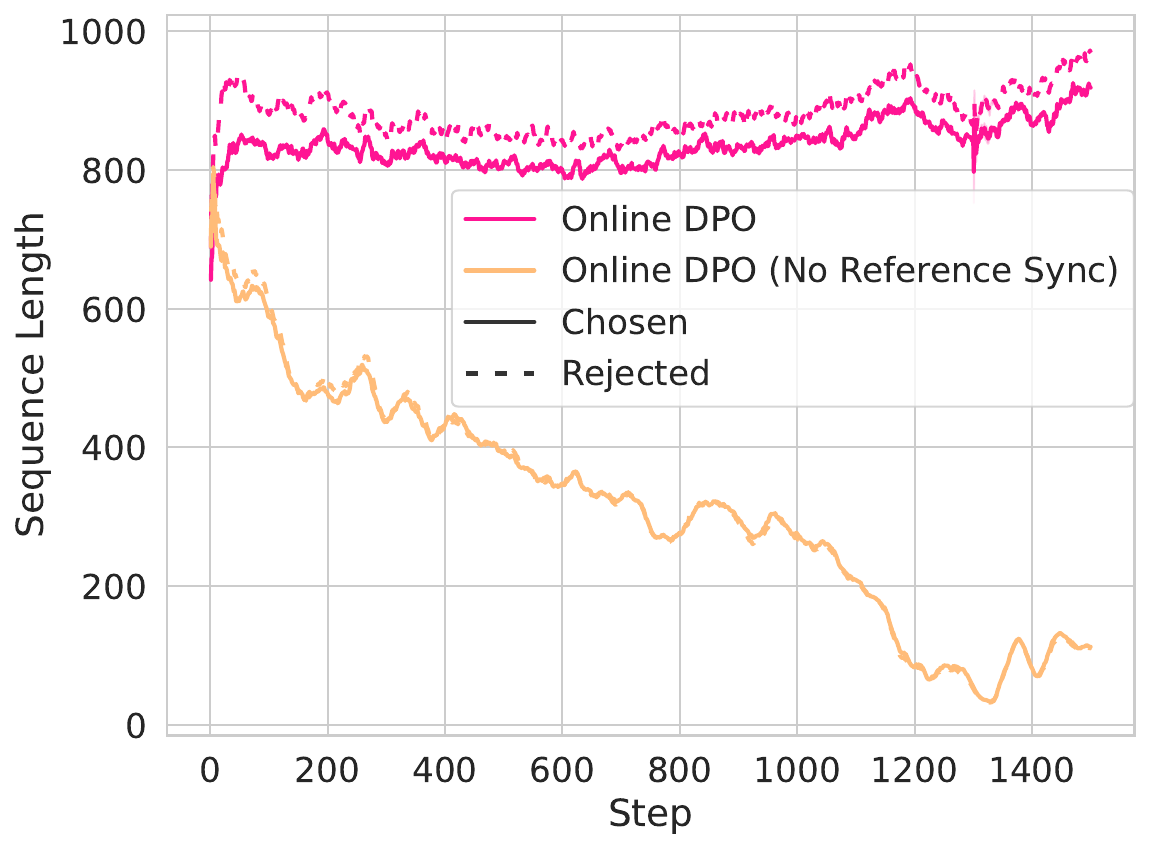}
        \caption{Lengths of the chosen versus rejected sequences in the training data for the non-collapsed (with reference model sync) and collapsed (without reference model sync) online DPO runs.}
    \end{subfigure}
    \caption{At the beginning of online DPO training, the model's shorter responses are more likely to be correct than longer responses (left). If training destabilizes (\emph{e.g.}, due to lack of reference model sync), the model reward hacks by producing excessively short sequences (right). However, if training remains stable, the model learns to gradually increase response length over time.}
    \label{fig:response_lengths}
\end{figure}

\begin{figure}[ht!]
    \centering
    \includegraphics[width=\linewidth]{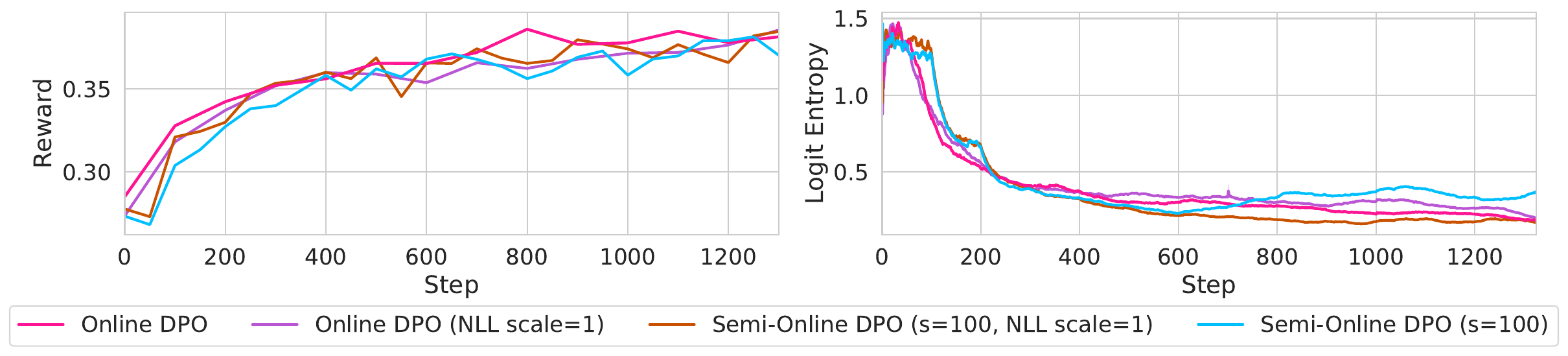}
    \caption{A comparison of online and semi-online DPO with and without an NLL term. Adding an NLL term does not provide benefits for either validation reward nor entropy in these settings.}
    \label{fig:dpo_nll}
\end{figure}

\newpage
\clearpage

\end{document}